%% file: root.tex
\documentclass{eptcs}
\usepackage{breakurl}             
\providecommand{\publicationstatus}{}

\usepackage{algorithm}
\usepackage{algorithmic}
\usepackage{graphicx} 
\usepackage{amsmath} 
\usepackage{amssymb}  
\usepackage{caption}
\usepackage{subcaption}
\usepackage{color}
\usepackage{tikz}
\usepackage[shortcuts]{extdash}

\graphicspath{{figures/}}

\usetikzlibrary{positioning, automata, arrows, shapes, fit, calc}

\title{
	Modeling Concurrency and Reconfiguration in Vehicular Systems: A $\pi$-calculus Approach
}

\author{Joseph Campbell \quad\ Cumhur Erkan Tuncali \quad Theodore P.~Pavlic \quad Georgios Fainekos
	\institute{School of Computing, Informatics, and Decision Systems Engineering\\Arizona State University, Tempe, AZ 85281, USA}
	\email{\{jacampb1, etuncali, tpavlic, fainekos\}@asu.edu}
	\thanks{This work has been partially supported by award NSF CPS 1446730.}
}

\def\titlerunning{Modeling Concurrency and Reconfiguration in Vehicular Systems: A $\pi$-calculus Approach}
\def\authorrunning{J.~Campbell, C.E.~Tuncali, T.P.~Pavlic \& G.~Fainekos}

\begin{document}

\maketitle
\thispagestyle{empty}
\pagestyle{empty}

\begin{abstract}

As autonomous or semi\-/autonomous vehicles are deployed on the roads, they will have to eventually start communicating with each other in order to achieve increased efficiency and safety.
Current approaches in the control of collaborative vehicles primarily consider homogeneous simplified vehicle dynamics and usually ignore any communication issues.
This raises an important question of how systems without the aforementioned limiting assumptions can be modeled, analyzed and certified for safe operation by both industry and governmental agencies.
In this work, we propose a modeling framework where communication and system reconfiguration is modeled through $\pi$-calculus expressions while the closed\-/loop control systems are modeled through hybrid automata.
We demonstrate how the framework can be utilized for modeling and simulation of platooning behaviors of heterogeneous vehicles.

\end{abstract}


\section{Introduction}

The DARPA Grand Challenges and, in particular, the Urban Challenge, demonstrated the feasibility of fully autonomous vehicles driving in urban and rural areas.
Since then, multiple well\-/established companies, research labs, and startups are competing towards becoming the first to sell fully autonomous vehicles capable of driving on the same roads as human\-/operated vehicles.
One important question that has not yet been addressed with current research and development activities is how to enable collaboration and cooperation among the autonomous (and even semi\-/autonomous) vehicles.

Cooperation is essential in many practical applications of autonomous vehicles, such as platooning. 
A vehicle platoon is a formation in which several vehicles closely follow each other in order to reduce aerodynamic drag, yielding both reduced fuel consumption~\cite{zabat1995aerodynamic} and increased road capacity~\cite{rao1993flow}. 
Another typical application where vehicle cooperation is required is collision avoidance at intersections~\cite{ColomboV12hscc}.
In this application, when two or more vehicles approach an intersection from different directions, they communicate directly or indirectly in order to cross the intersection without (usually) coming to a full stop.

Prior work has tackled the challenge of vehicle cooperation by focusing on the development of control and scheduling algorithms for homogeneous vehicles with simplified dynamics.
Another common assumption in the current literature is to ignore any potential issues in the communication protocols of the vehicular systems.
Both assumptions can be very limiting when trying to utilize such control algorithms on real networks of heterogeneous vehicles operating at high speeds.

In this paper, we develop a modeling framework for collaborating vehicular systems where both the high\-/level communication protocols and the low\-/level complex vehicle dynamics can be modeled in the same framework.
For modeling the communication and reconfiguration layer of the system we have chosen the formalism of $\pi$-calculus, which was specifically developed for modeling and reasoning over mobile communicating processes where the network structure can be modified dynamically~\cite{milner1999communicating}.
For closed\-/loop system dynamics of each vehicle, we use the modeling framework of hybrid automata~\cite{alur2015principles}.
The two layers communicate and synchronize through message passing.

The primary and immediate benefit of such a framework is its flexibility.
Different communication protocols can be modeled quickly in a hierarchical fashion where at the highest level we can verify correctness of the protocols. 
At the same time, executable code can be automatically generated for the lower levels.
For example, the proposed framework can easily incorporate verifiable privacy protocols~\cite{DelauneRS2008} or broadcasting protocols~\cite{singh2010process}.
Similarly, different vehicles and control algorithms can be easily modeled and simulated because hybrid automata model both the control and continuous dynamics of the vehicles as well as any other discrete modes of operation for these vehicles~(e.g., emergency braking, economy versus sport driving mode).
The main benefit of utilizing hybrid automata as a modeling framework is that reachability analysis~\cite{FrehseCAV11} and automated test generation~\cite{AnnapureddyLFS11tacas} can be performed efficiently and effectively.

Because autonomous vehicles are safety\-/critical systems, the ultimate goal of this project is the requirements\-/based analysis of the whole system rather than of specific components~(e.g.,~\cite{singh2010process}) or behaviors~(e.g.,~\cite{AlthoffEtAl2010ivs}).
Namely, we envision a verification framework for heterogeneous, high\-/fidelity models of collaborating vehicular systems similar to what was proposed by Damm et~al.~\cite{Damm2007} for the Automatic Train Protection system.
In particular, stable and safe vehicle controllers for simplified models can be utilized in high\-/fidelity vehicle models in order to be further analyzed using automated test generation tools such as \textsc{S-TaLiRo}~\cite{AnnapureddyLFS11tacas}.

In summary, our contribution in this paper is the development of a modeling framework that can represent complex vehicular networks. 
The complexity enters both in terms of vehicle dynamics and in terms of complex communication and decision\-/making protocols.
We show that our framework can effectively model heterogeneous vehicular networks running a variety of control algorithms.
Finally, the proposed framework lays the groundwork in making verification of these systems both tractable and practical.

\section{Related Work}
\label{sec:related}

A great deal of effort has gone into proving the safety of autonomous vehicle systems. 
Due to the large and diverse literature, we will present a few works that have inspired our own approach.
The reader can find further references in the literature discussed below.

Some methods, such as those by Asplund et~al.~\cite{asplund2012formal}, use formal methods to verify cooperation among autonomous vehicles.
In particular, the authors formalize a distributed coordination protocol using finite state machines, and they consider an over\-/approximation of the vehicles' dynamics based on the maximum acceleration of the vehicles. 
Then, they utilize Satisfiability Modulo Theory~(SMT) solvers to prove safety of an intersection collision avoidance protocol.
The results cannot be generalized since they apply only to scenarios where the intersection is a shared resource where only one vehicle has access at each time.

Lygeros et~al.~\cite{lygeros1998verified} model platooning as a hybrid automaton, and they automatically design control laws for provably safe merge and split vehicle operations.
However, their work does not look into the formalization of the coordination layer for the vehicles and instead treats it rather informally.
Our work can be thought of as enabling the modeling of more complicated communication and control algorithms on top of or as an extension of the hybrid automaton models proposed by Lygeros et~al.~Beyond more complex coordination protocols that include privacy and security, we can study for example the behavior of the platoons when there is an unexpected obstacle on the road.

More recently, Loos et~al.~\cite{loos2011adaptive} apply theorem proving methods to verify the safety of adaptive cruise controllers.
In particular, they model the case of vehicles with known bounds on important characteristics (i.e., maximum braking deceleration, maximum acceleration, and worst\-/case response\-/time) with cruise\-/control implementations which pick an acceleration subject to a special safety condition.
Under that condition, they show that the resulting arbitrarily large group of vehicles will be collision free.
A similar result is shown by Platzer~\cite{Platzer12} for the case of vehicles entering a highway from an on\-/ramp, where collision\-/free safety can be proved under the assumption that vehicles entering the highway meet a condition based on their position, speed, and acceleration relative to vehicles on the highway.
Although it is hinted that this framework can capture the effect of a communication\-/mediated reconfiguration of the vehicle platoon, neither case explicitly models any communication nor coordination protocols between vehicles.
Platzer~\cite{Platzer12} clarifies that communication can be modeled within their framework, but shared variables with access delays must be used as opposed to explicitly modeling communication channels that are opened, maintained, and closed by vehicles.
Furthermore, despite the expressiveness of the framework, direct and automated synthesis of controllers that meet required specifications remains elusive.

Along another direction to this problem, Franzle et~al.~\cite{franzle2015no} present an extension to the Multi\-/Lane Spatial Logic~(MLSL)~\cite{hilscher2011abstract} by introducing a local scope for the observations of each vehicle. 
MLSL has been proposed as a logical framework for reasoning about decision\-/making algorithms for automated driving.

Finally, a number of works consider the verification and synthesis problem for cooperative vehicle behaviors exclusively at the supervisory or communication level only. 
That is, no continuous vehicle dynamics are explicitly considered.
For example, Bochmann et~al.~\cite{bochmann2015synthesizing} present a discrete\-/event controller synthesis algorithm for lane\-/changing maneuvers.
Bengtsson et~al.~\cite{bengtsson2015interaction} present a communication protocol for enabling two platoons to merge together.
The work of Kamali et~al.~\cite{kamali2016formal} is very similar to our proposed architecture in the sense that they also enable simulation of collaborative vehicles by utilizing high\-/fidelity models at the physical level.
However, the vehicle coordination protocol is modeled through timed automata instead of mobile process calculi.

\section{Preliminaries}
\label{sec:prelim}
In this paper, we model concurrent high\-/level behavior of autonomous vehicles with a process algebra known as $\pi$-calculus. This section provides a brief overview of the algebra along with an extension that is necessary to fully represent a distributed autonomous system.

\subsection{Process Algebra}

Process algebras, or process calculi as they are also known~\cite{baeten2005brief}, are a way to algebraically model the interaction between concurrent systems. They are exceedingly useful for reasoning about parallel systems and as such allow for the verification of these systems to determine whether certain properties hold. For example, if we are considering a system of autonomous vehicles that are traveling together in a platoon, then we would like to reason about such a system and establish that merging behavior protocols never produce a deadlock.

To this end, we must first establish the notion of a process. Informally speaking, a process is simply the behavior of a system~\cite{baeten2005brief}. Process algebra puts forward the idea that these behaviors can be modeled as a sequence of actions over time which can be manipulated with algebraic rules. As a simple example, consider an abstracted model of a process for a vehicle that is changing lanes. Given the actions, $enable\_signal$, $disable\_signal$, $change\_lane$, and the sequential operator ($.$) this process can be modeled as $enable\_signal.change\_lane.disable\_signal$. This expression means that the observed behavior is to enable the turn signal, change lanes, and finally disable the turn signal. However, as we describe in more detail in the following section, the real power of process algebra comes from its ability to describe the interaction between concurrent systems.

\subsection{$\pi$-Calculus}

In this work, we employ a specific variant of process algebra known as $\pi$-calculus~\cite{milner1992calculus, milner1999communicating}. In its most basic form, $\pi$-calculus defines an expression for a process $P \in \mathcal{P}$ as a sequence atomic actions $\pi$. The syntax of a process expression can be described by the following grammar.
\begin{subequations}
\begin{equation}
P ::= \pi . P \mid 0
\end{equation}
Semantically, this means the process $P$ is empty (0) and thus
terminating or is composed of one or more atomic actions $\pi$ that
execute in sequence and yield another process expression. Atomic actions
in $\pi$-calculus are made up of primitives known as \emph{names}, which
may be variables, silent actions, communication channels, or simply data
values; there is no distinction between them. Given an infinite set of
names $\mathcal{N}$ and $w, x, y, z \in \mathcal{N}$, three types of
actions are defined. Namely,
\begin{itemize}
	\item Unobservable actions occur when the effect of execution is unknown to the $\pi$-calculus process. For example, $P = w.x.P'$ defines $P$ as executing $w$ and $x$ sequentially and in order, which results in the process $P'$. Note that this resulting process can also be $P$ itself for recursion, as in $P = w.x.P$. Alternatively, if the process terminates after the execution of $w$ and $x$ then it will be defined as $P = w.x.0$. For the sake of brevity, a terminating process typically excludes the $.0$ postfix.
	\item Outgoing communication occurs when a name is sent along out over a channel. The expression $P = \overline{x}{<}y{>}$ defines $P$ as sending the name $y$ out over $x$. In this example, $x$ represents a communication channel to another process, and $P$ is sending the message $y$ over that channel.
	\item Incoming communication occurs when a name is received along a channel. The expression $P = x(z)$ defines $P$ as receiving the name $z$ from $x$. Similar to the previous example, $x$ represents a communication channel to another process, and $P$ receives a message on that channel which gets bound to the name $z$.
\end{itemize}
In addition, process expressions may take a parametric form. For example, $P(x) = \overline{x}{<}y{>}$ is a parametric definition of process $P$. When $P$ is referenced, it must be passed a name. If we have the expression $w(z).P(z)$, a message is received over channel $w$ that gets bound to $z$ and is then passed to $P$. In this context, $z$ must be another communication channel itself because $P$ immediately uses it to send message $y$. This ability to transmit and receive communication channels is an important concept in $\pi$-calculus and is referred to as \emph{mobility}.

We now introduce more complex constructs into the syntax, as summarized by the following grammar.
\begin{gather}
P ::= P \mid P || P' \mid P + P' \mid {!}P'\\
P ::= P \mid (\nu x)P \mid x:[y \Rightarrow P, z \Rightarrow P']
\end{gather}
\end{subequations}
These constructs support the modeling of multiple processes. In particular,
\begin{itemize}
	\item $P = P || P'$ defines the process $P$ as concurrently executing $P$ and $P'$.
	\item $P = P + P'$ defines the process $P$ as non\-/deterministically executing either $P$ or $P'$.
	\item $P = !P'$ indicates that there are (potentially) an infinite number of copies of $P'$ executing concurrently. This is equivalent to $P = P' || P$. For an actual system, an infinite number of copies is not possible, and so this represents as many concurrently executing copies as desired.
	\item $(\nu x)P$ indicates the name $x$ is unique to $P$.
    \item $P = x:[y \Rightarrow P, z \Rightarrow P']$ defines the process $P$ as resulting in $P$ if the name $x$ is equal to $y$, or resulting in $P'$ if $x$ is equal to $z$. Names can also be constants, and so it is possible define $P = x:[\text{True} \Rightarrow P, \text{False} \Rightarrow P']$ for evaluating against True and False.
\end{itemize}

Now that the syntax and an informal explanation of the semantics for concurrent processes have been introduced, the concept of \emph{reaction} must be discussed. Just as there are unobservable actions, there are also observable actions. Consider two processes, $P = w.P'$ and $Q = \overline{w}.Q'$, and suppose they are running concurrently, $P||Q$. The names $w$ and $\overline{w}$ are considered complementary, and the actions are linked; if $\overline{w}$ is executed, then $w$ will also be executed. This is why we say the actions are observable: the effect of executing $\overline{w}$ is observed by the $\pi$-calculus rules as it causes $w$ to also execute. The similarity in syntax between observable actions and communication actions is no coincidence, as communication actions are observable actions. For a formal analysis of reactions, refer to Milner~\cite{milner1999communicating}.

This highlights the importance of the scope restriction operator, $\nu$. Let $P(x) = \overline{x}{<}y{>}$, $Q(x) = x(z)$, and $R(x) = x(w)$. If the processes run concurrently as in $P(x) || Q(x) || R(x)$ then there are two outcomes: either $Q$ receives the message sent by $P$ or $R$ does. This is a consequence of all three processes sharing the same $x$ channel. However, if $x$ is now restricted as in $(\nu x)(P(x) || Q(x)) || R(x)$ then there is only one outcome: $Q$ receives the message sent by $P$ along their shared (and unique) channel $x$, and $R$ will receive another message on a different channel $x$ from an unobserved process.

\subsection{$\omega$-Calculus}
This work also makes use of $\omega$-calculus, an extension to $\pi$-calculus that introduces semantics for reasoning about mobile ad hoc wireless networks~\cite{singh2010process}. It is a particularly useful extension when dealing with distributed robotic systems that must form ad hoc networks in order to establish communication. In this work, two particular operations from $\omega$-calculus are important:
\begin{itemize}
	\item $\overline{\textbf{b}}{<}x{>}$ broadcasts the message $x$ to any recipient in transmission range.
	\item $\textbf{r}(y)$ receives a message that has been broadcasted and binds it to $y$.
\end{itemize}

\section{Methodology} \label{sec:methodology}

\begin{figure}
	\centering
	\includegraphics[width=.5\linewidth]{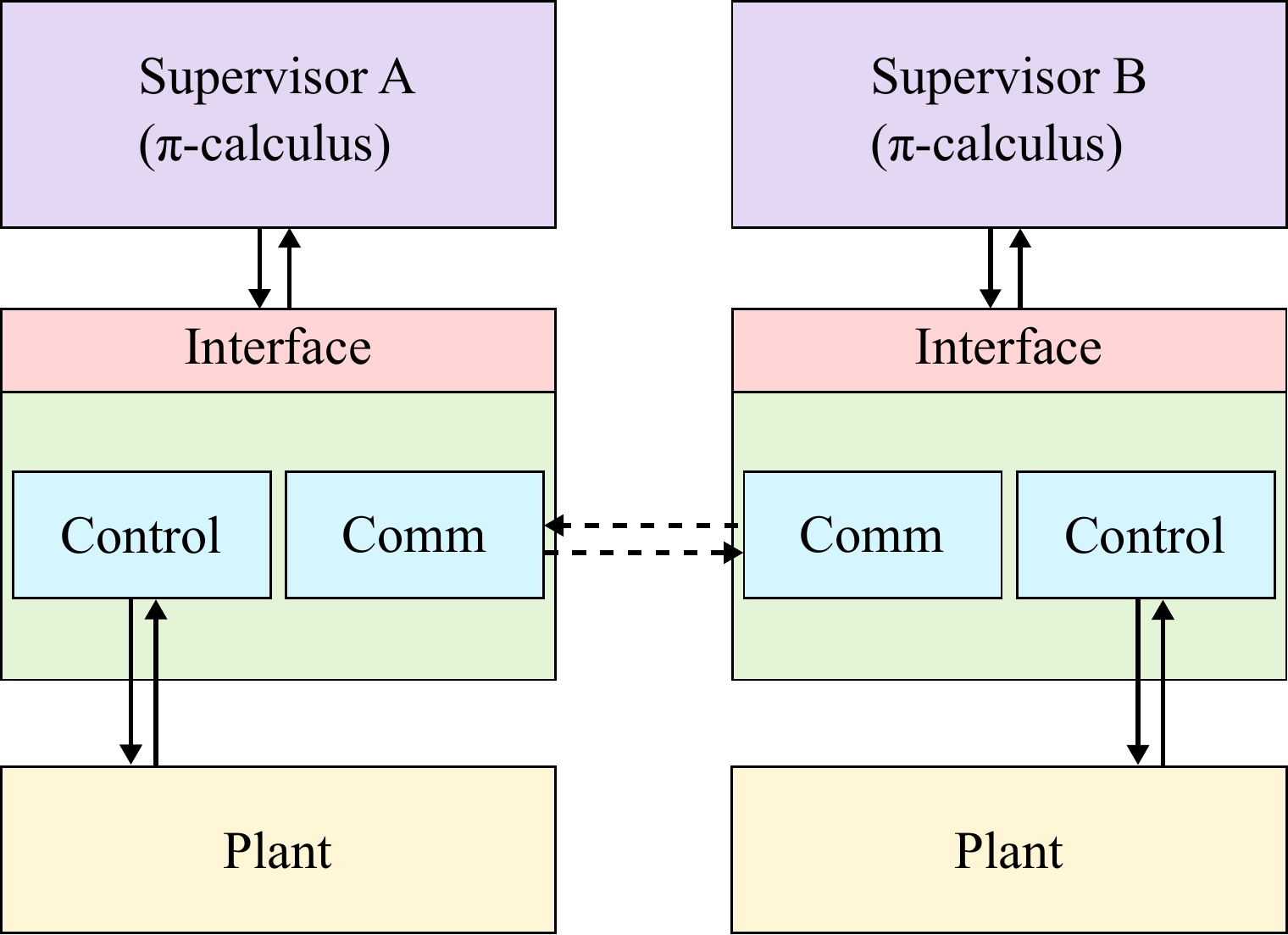}
	\caption{Organization of two concurrent vehicle processes.}
	\label{fig:org_diagram}
\end{figure}

As previously discussed, autonomous vehicle platoons are a tantalizing goal due to the inherent benefits they bring in the form of increased road capacity and reduced energy usage~\cite{zabat1995aerodynamic, rao1993flow}. However, designing such a system is not a simple endeavor. In order to scale effectively over a large number of vehicles, this distributed platoon system should have decentralized control. Additionally, the benefits and stability of a platoon are increased if the vehicles remain in communication with each other~\cite{lygeros1998verified}. As autonomous vehicles are safety\-/critical systems, the behavior that governs any such vehicle platoon must be verified to ensure no unforeseen dangerous scenarios can arise. Further complicating matters, autonomous vehicles are not homogeneous systems; each vehicle will have varying physical characteristics. Our work tackles these issues by dividing the problem into two non\-/overlapping parts: a high\-/level discrete\-/logic layer that describes the overall behavior of the autonomous vehicles, and a low\-/level physical layer that describes the closed\-/loop dynamics of the underlying system. Interaction occurs in the form of an interface provided by the low\-/level layer through which the high\-/level layer can send signals. This organization is shown in Fig.~\ref{fig:org_diagram}. The following sections describe these layers in more detail.

\subsection{High-level Layer} \label{ssec:highlevel}

The high\-/level layer must be capable of satisfying three conditions: it must be able to model the protocols of a distributed multi\-/vehicle system, it must be able to model communication between vehicles, and it must be amenable to verification processes for safety\-/critical systems. As it turns out, $\pi$-calculus meets these three conditions and is used to model the high\-/level behavior in this work. First, however, we must establish the expected platoon behavior.

\begin{figure}
	\centering
	\begin{subfigure}{.27\linewidth}
		\centering
		\includegraphics[width=.6\linewidth]{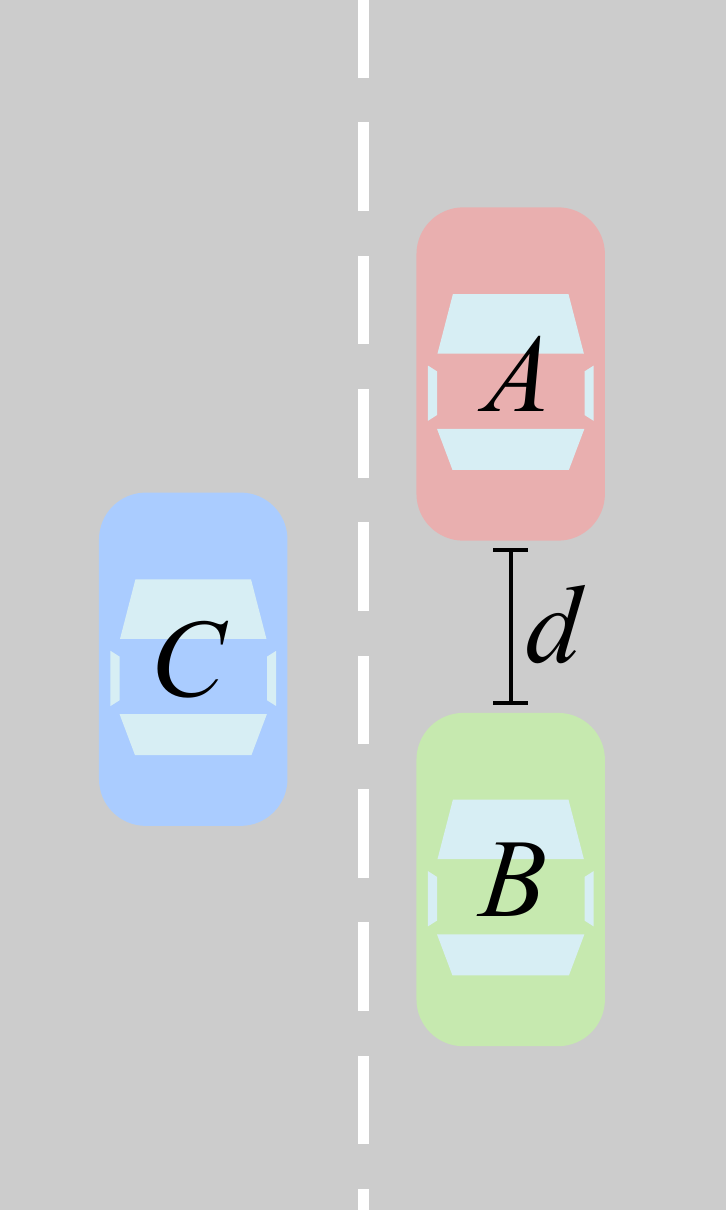}
		\caption{Initial $A$\--$B$ spacing}
		\label{fig:convoy_diagram1}
	\end{subfigure}%
	\begin{subfigure}{.27\linewidth}
		\centering
		\includegraphics[width=.6\linewidth]{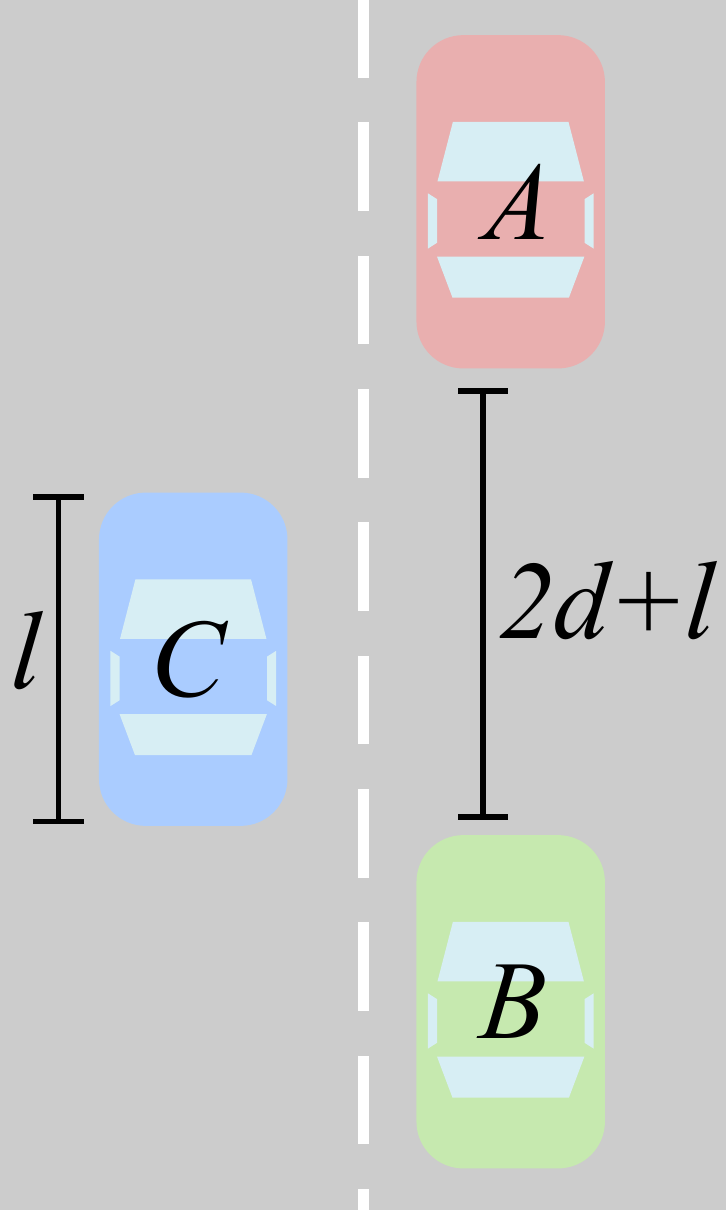}
		\caption{After space is created for $C$}
		\label{fig:convoy_diagram2}
	\end{subfigure}
	\caption{Expected spacing behavior of vehicle convoy.}
	\label{fig:convoy_diagram}
\end{figure}

Figure~\ref{fig:convoy_diagram1} depicts a simple platoon of two vehicles. The vehicle labeled $A$ is the leader of this platoon and the vehicle labeled $B$ is following $A$ at a minimum safe distance $d$. A third vehicle, $C$, is outside of the platoon in a separate lane. Suppose $C$ wishes to join the platoon and that the optimal place to do so is between $A$ and $B$. Before $C$ is permitted to join, the distance between $A$ and $B$ must be increased to $2d + l$, where $l$ is the length of $C$ as shown in Fig.~\ref{fig:convoy_diagram2}. This action ensures that the minimum safe distance is always respected. This behavior can be further simplified by letting $C$ follow $A$ at distance of $d$ while $B$ also follows $C$ at a distance of $d$, making explicit knowledge of $l$ unnecessary.

In this work, we define separate behaviors for three different scenarios: a leader of a platoon~(Leader), a follower in a platoon~(Follower), and a vehicle joining a platoon~(Joiner). These are the behaviors exhibited by vehicles $A$, $B$, and $C$, respectively, in Fig.~\ref{fig:convoy_diagram}. It is also possible for vehicles to transition between behaviors. For example, a vehicle joining a platoon will become a Follower once it has fully merged. At the highest level, the system modeled in Fig.~\ref{fig:convoy_diagram} can be represented as a $\pi$-calculus expression. Let $A = Leader$, $B = Follower$, and $C = Joiner$. This system is then described by $A || B || C$. Each of these behaviors will now be described in terms of a $\pi$-calculus expression.

In Section~\ref{sec:methodology}, it was stated that an interface layer is utilized by the behaviors to exert control over the underlying continuous system. The interface layer represents the unobservable actions that can be called by a $\pi$-calculus process expression. Although these actions are considered atomic to $\pi$-calculus expressions, this will not be the case with respect to the low\-/level continuous layer. In particular, some actions take considerable time to execute and may wait until certain conditions unknown to the high\-/level layer are satisfied. The following unobservable actions are provided in the interface layer.
\begin{itemize}
	\item $get\_id$ gets the identification of the vehicle. If $B$ in Fig.~\ref{fig:convoy_diagram1} calls $get\_id$, it will return $B$.
	\item $get\_ldr$ gets the leader of the vehicle. As an example, if $B$ in Fig.~\ref{fig:convoy_diagram1} calls $get\_ldr$, it will return $A$.
	\item $set\_ldr$ sets the leader of the vehicle to the vehicle with the given identification.
	\item $drive$ tells the vehicle to drive forward with respect to the road geometry.
	\item $keep\_dist$ maintains a safe following distance $d$ from the current leader.
	\item $check\_join$ checks whether the vehicle with the given identification occupies the position where this vehicle wants to join. For example, if $C$ in Fig.~\ref{fig:convoy_diagram1} invokes $check\_join$ on $B$, it will return positive.
	\item $align\_start$ causes the interface to trigger the $align\_done$ event action when the vehicle is $d$ distance from the current leader.
	\item $merge\_start$ causes the interface to trigger the $merge\_done$ event action when the vehicle has finished merging into the current leader's lane.
\end{itemize}
Additionally, recursive expressions are only allowed to execute at most once per sampling period.

\begin{figure}
	\centering
	\begin{subfigure}{.22\linewidth}
		\centering
		\includegraphics[width=.6\linewidth]{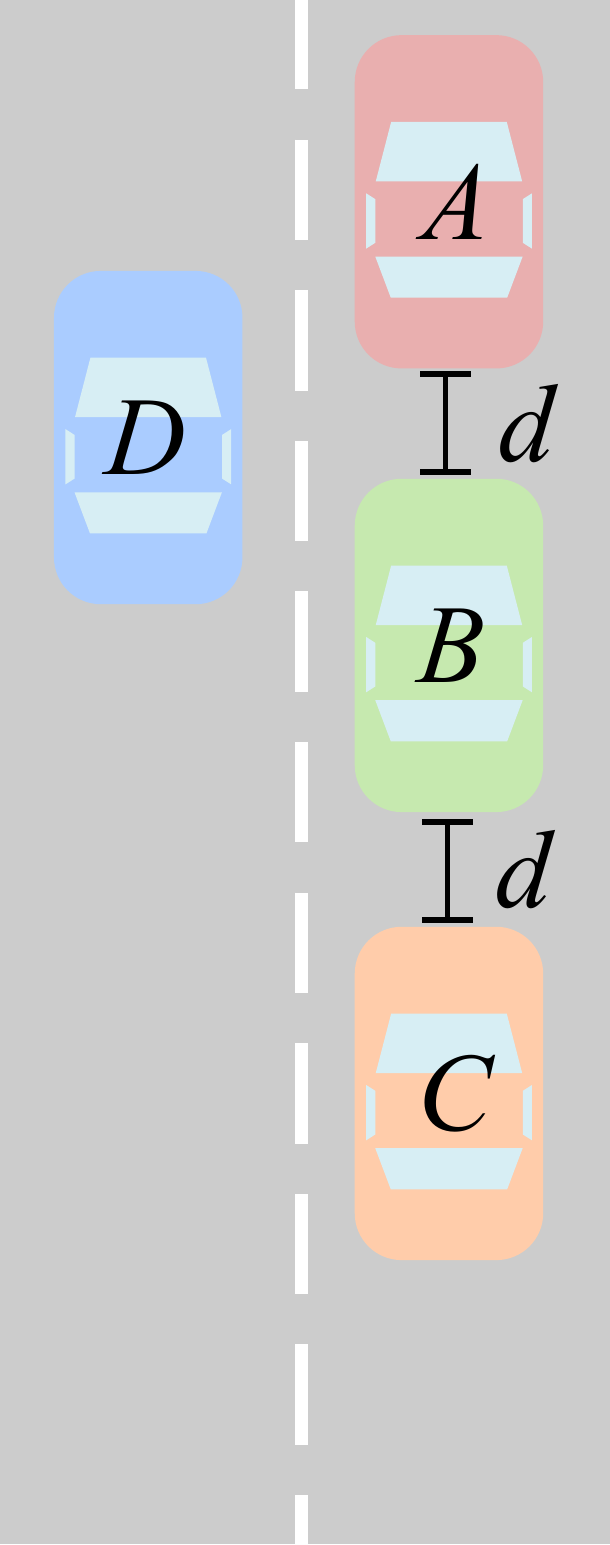}
		\caption{Initial state}
		\label{fig:convoy_merge1}
	\end{subfigure}%
	\begin{subfigure}{.22\linewidth}
		\centering
		\includegraphics[width=.6\linewidth]{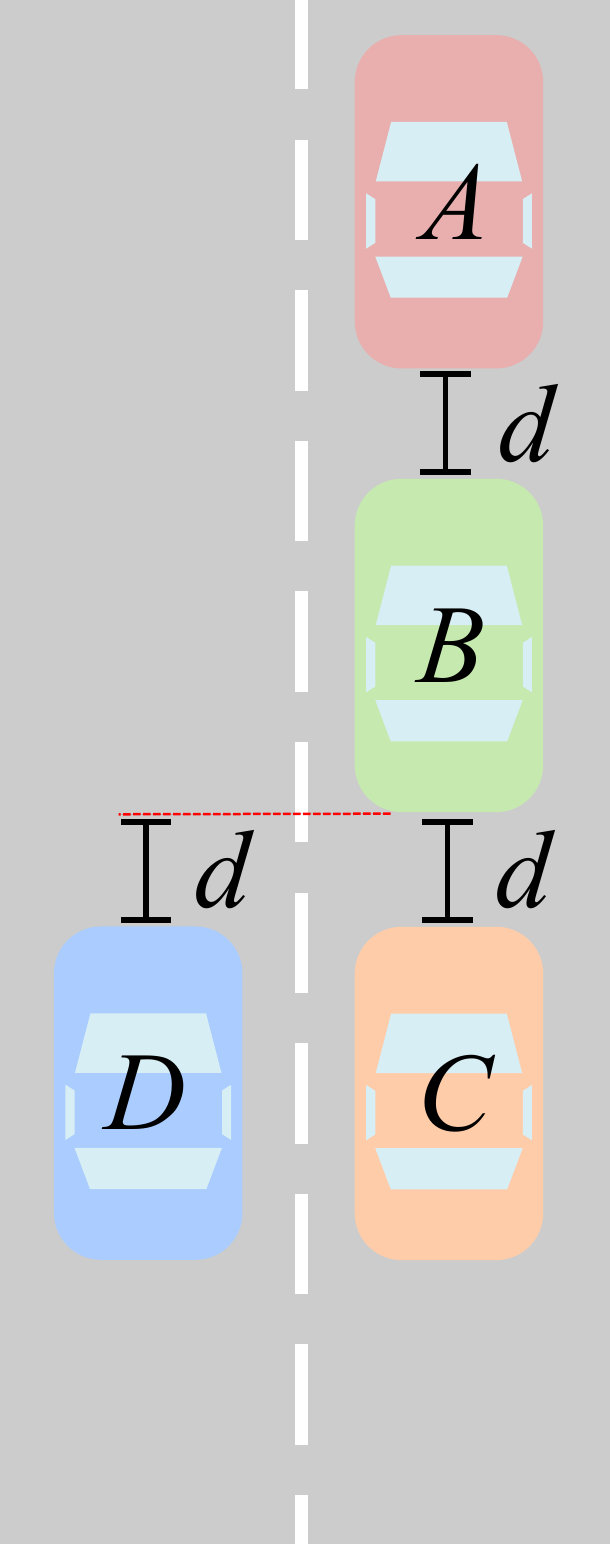}
		\caption{Joiner D in position}
		\label{fig:convoy_merge2}
	\end{subfigure}%
	\begin{subfigure}{.22\linewidth}
		\centering
		\includegraphics[width=.6\linewidth]{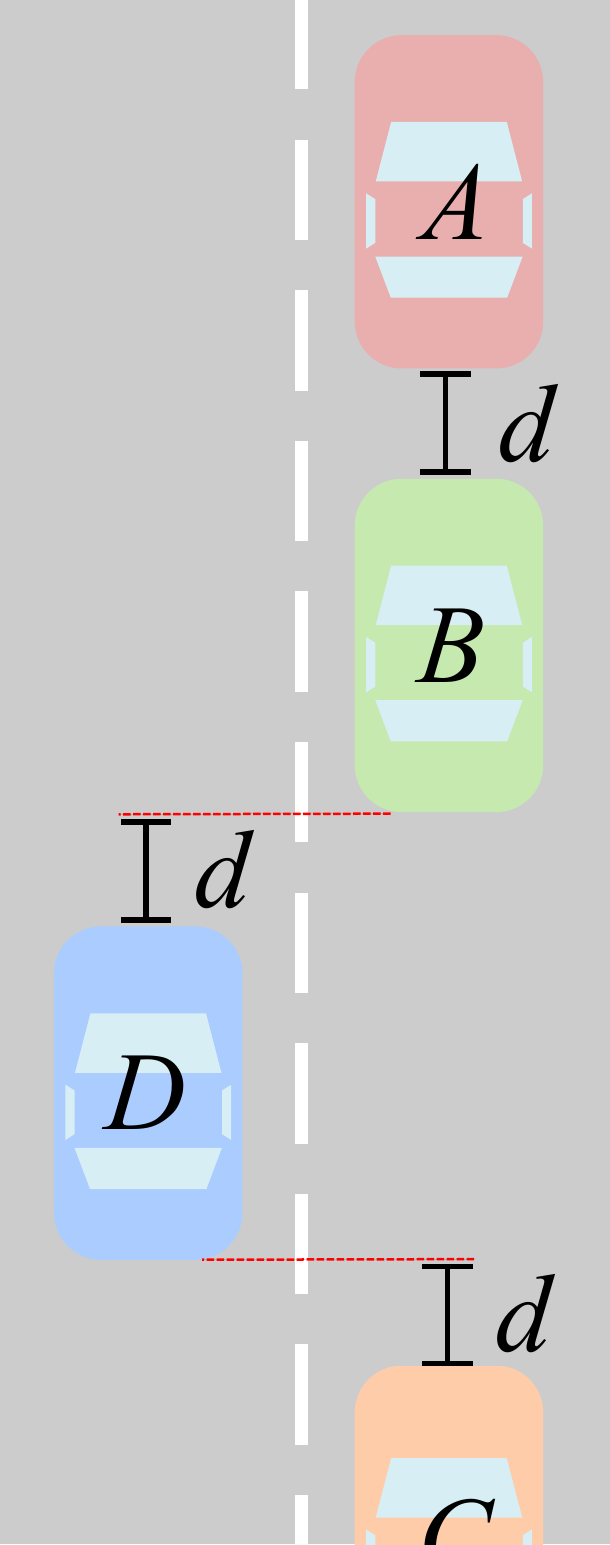}
		\caption{Distance created}
		\label{fig:convoy_merge3}
	\end{subfigure}%
	\begin{subfigure}{.22\linewidth}
		\centering
		\includegraphics[width=.6\linewidth]{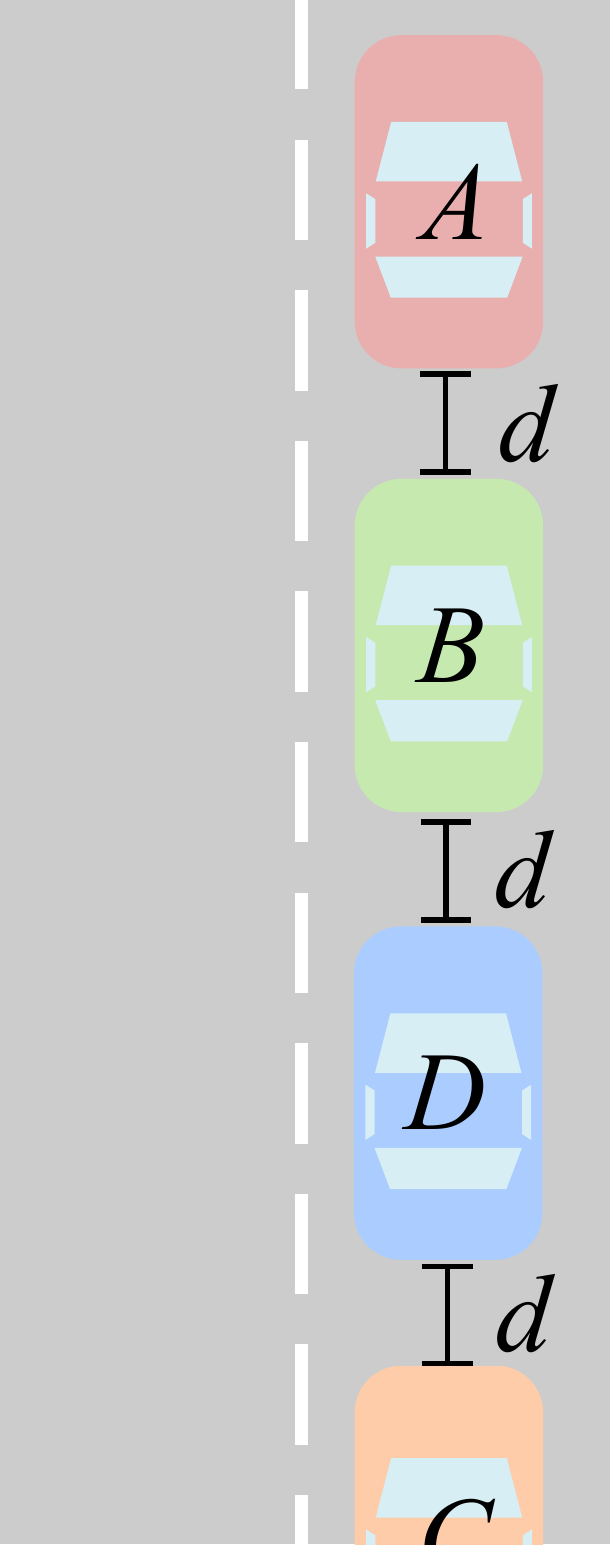}
		\caption{Merge complete}
		\label{fig:convoy_merge4}
	\end{subfigure}
	\caption{Expected merge behavior of vehicle platoon.}
	\label{fig:convoy_merge}
\end{figure}

The behavior descriptions that follow refer to the system depicted in Fig.~\ref{fig:convoy_merge}.
$A$, $B$, and $C$ are initially in a platoon of which $A$ is the Leader and $B$ and $C$ are Followers. $D$ is a Joiner external to the platoon.

\subsubsection{Leader} \label{sssec:leader}
The behavior of a platoon Leader is straightforward in this scenario: it simply drives forward indefinitely. The associated process expression in Algorithm~\ref{alg:leader_rule} invokes the atomic action $drive$ and repeats this action indefinitely.
\begin{algorithm}
	\caption{Leader}
	\label{alg:leader_rule}
	\begin{algorithmic}[1]
		\STATE Leader$ = \overline{drive} . \text{Leader}$
	\end{algorithmic}
\end{algorithm}

\subsubsection{Follower} \label{sssec:follower}

The behavior of a Follower in a platoon is much more involved as it must simultaneously follow its leader while cooperating with vehicles who wish to join the platoon. This is modeled by the concurrent execution of Follow and Cooperate in the Follower state in Algorithm~\ref{alg:follower_rule}, which corresponds with the initial state in Fig.~\ref{fig:convoy_merge1}. Just as Leader in Algorithm~\ref{alg:leader_rule} is an infinitely repeating call to the atomic action $drive$, Follow is an indefinitely repeating call to the atomic action $keep\_dist$.
\begin{algorithm}
	\caption{Follower}
	\label{alg:follower_rule}
	\begin{algorithmic}[1]
		\STATE Wait$(y) = y . \overline{merge\_done}$
		\STATE Align$(y) = \overline{align\_start} . align\_done . \overline{y} . \text{Wait}$
		\STATE Rcv\_Ldr$(y,ldr) = y(nldr) . \overline{set\_ldr}{<}nldr{>} . \text{Align}(y)$
		\STATE Send\_Ldr$(y) = get\_ldr(ldr) . \overline{y}{<}ldr{>} . \text{Rcv\_Ldr}(y,ldr)$
		\STATE Respond$(y, flag) = flag:[True \Rightarrow \text{Send\_Ldr}(y)]$
		\STATE Ident$(y) = get\_id(id) . \overline{y}{<}id{>} . y(flag) . \text{Respond}(y, flag)$
		\STATE Cooperate$ = !\textbf{r}(x) . (\nu y)(\overline{x}{<}y{>} . \text{Ident}(y))$
		\STATE Follow$ = \overline{keep\_dist} . \text{Follow}$
		\STATE Follower$ = \text{Follow} || \text{Cooperate}$
	\end{algorithmic}
\end{algorithm}

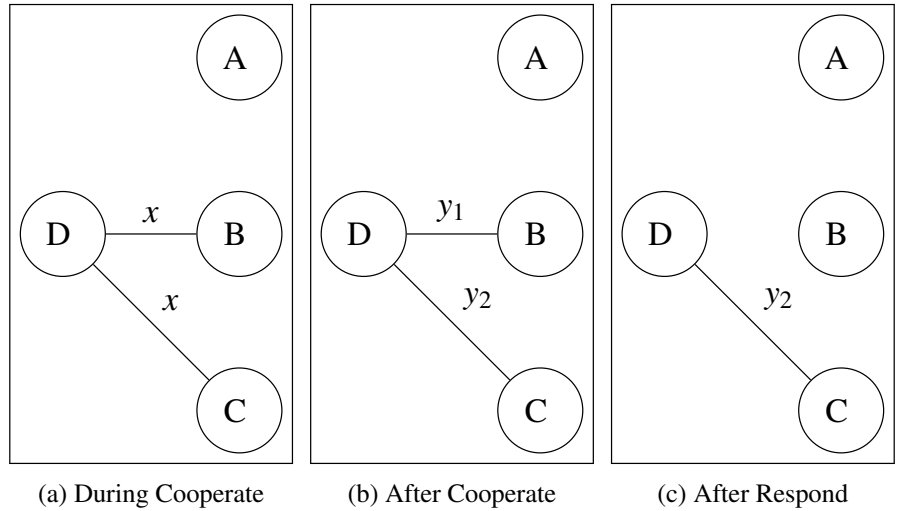
\begin{figure}
	\centering
	\begin{subfigure}{.25\linewidth}
		\centering
		\resizebox{1.0\linewidth}{!}{\input{"figures/convoy_mobility1.tex"}}
		\caption{During Cooperate}
		\label{fig:convoy_mobility1}
	\end{subfigure}%
	\begin{subfigure}{.25\linewidth}
		\centering
		\resizebox{1.0\linewidth}{!}{\input{"figures/convoy_mobility2.tex"}}
		\caption{After Cooperate}
		\label{fig:convoy_mobility2}
	\end{subfigure}%
	\begin{subfigure}{.25\linewidth}
		\centering
		\resizebox{1.0\linewidth}{!}{\input{"figures/convoy_mobility3.tex"}}
		\caption{After Respond}
		\label{fig:convoy_mobility3}
	\end{subfigure}
	\caption{Process graph depicting communication network during various points of execution. Cooperate and Respond refer to Lines 7 and 5 of Algorithm~\ref{alg:follower_rule} respectively. A, B, C, and D correspond to the vehicles in Fig.~\ref{fig:convoy_merge}.}
	\label{fig:convoy_mobility}
\end{figure}

Cooperate (Line 7 in Algorithm~\ref{alg:follower_rule}) makes use of $\pi$-calculus's special notion of \emph{mobility}, which allows us to reconfigure the communication network between vehicles and reason about it. When a vehicle wishes to join a platoon, it creates a communication channel $x$ and broadcasts it to every vehicle within range in Line 8 of Algorithm~\ref{alg:join_rule}. This establishes a communication channel between the Joiner and every Follower as shown in Fig.~\ref{fig:convoy_mobility1}. However, a unique communication channel is desired between the Joiner and each Follower, so the first thing a Follower does upon receiving channel $x$ is to create a new unique channel $y$ and send that to the Joiner in Line 7 of Algorithm~\ref{alg:follower_rule}. This results in a reconfiguration of the network as shown in Fig.~\ref{fig:convoy_mobility2}. Eventually, the Joiner will decide where to merge in the platoon and will drop all communication channels except to the vehicle that will follow the Joiner, resulting in the final network shown in Fig.~\ref{fig:convoy_mobility3}. It is possible to expand these interactions and establish a multi-hop communication network between a vehicle and its immediate peers when it joins the platoon, however, this is not examined in this work.

Once a unique communication channel is established between the joining vehicle and each Follower as in Fig.~\ref{fig:convoy_mobility2}, it is used to synchronize the joining process. In Line 6 of Algorithm~\ref{alg:follower_rule}, the Follower transmits its own identifier to the joining vehicle via $y$ so that it can decide whether to join at the Follower's position. Once a reply is received and bound to $flag$, the Respond state (Line 5) will either continue to Send\_Ldr if $flag$ is True or terminate otherwise. If $flag$ is True, the joining vehicle intends to join the platoon at the Follower's location. Before this can happen, the joining vehicle must position itself next to the Follower so that it is ready to merge over as shown in Fig.~\ref{fig:convoy_merge2}. In order to facilitate this, the Follower transmits its current leader, $ldr$, to the joining vehicle.

When the joining vehicle is in position, it responds to the Follower with its own identifier, $nldr$, so that the Follower can set the joining vehicle as its new leader in the Rcv\_Ldr state. After setting the joining vehicle as its new leader, the Follower increases its distance to $d$ from the joining vehicle due to the concurrently executing Follow state. In the Wait state, the Follower waits until the distance between it and the joining vehicle reaches $d$ with the atomic action $wait\_dist$. This is depicted in Fig.~\ref{fig:convoy_merge3}.
Lastly, the joining vehicle is signaled to merge, which results in Fig.~\ref{fig:convoy_merge4}.

\subsubsection{Joiner} \label{sssec:joiner}

The behavior of a Joiner is given in Algorithm~\ref{alg:join_rule}. Until a decision to join is made by the high\-/level layer, the joining vehicle drives forward with respect to the road geometry just like a Leader. The one novel detail beyond the previous description is that the Joiner behavior transitions to a Follower as the last action in the Merge state.
\begin{algorithm}
	\caption{Joiner}
	\label{alg:join_rule}
	\begin{algorithmic}[1]
		\STATE Merge$(y) = \overline{merge\_start} . merge\_done . \overline{y} . \text{Follower}$
		\STATE Wait$(y) =  get\_id(id) . \overline{y}{<}id{>} . y . \text{Merge}$
		\STATE Align$(y) = \overline{align\_start} . align\_done . \text{Wait}(y)$
		\STATE Rcv\_Ldr$(y) = y(ldr) . set\_ldr(ldr) . \text{Align}(y)$
		\STATE Ans$(y,ok) = \overline{y}{<}ok{>} . ok:[T \Rightarrow \text{Rcv\_Ldr}(y)]$
		\STATE Check$(y,id) = (\nu z)(\overline{join\_ok}{<}z,id{>} . z(ok) . \text{Ans}(y,ok))$
		\STATE Listen$(x) = x(y) . y(id) . \text{Check}(y,id)$
		\STATE Joiner$ = (\nu x)(\overline{\textbf{b}}{<}x{>} || !\text{Listen}(x))$
	\end{algorithmic}
\end{algorithm}

\subsection{Low-level Layer} \label{ssec:lowlevel}
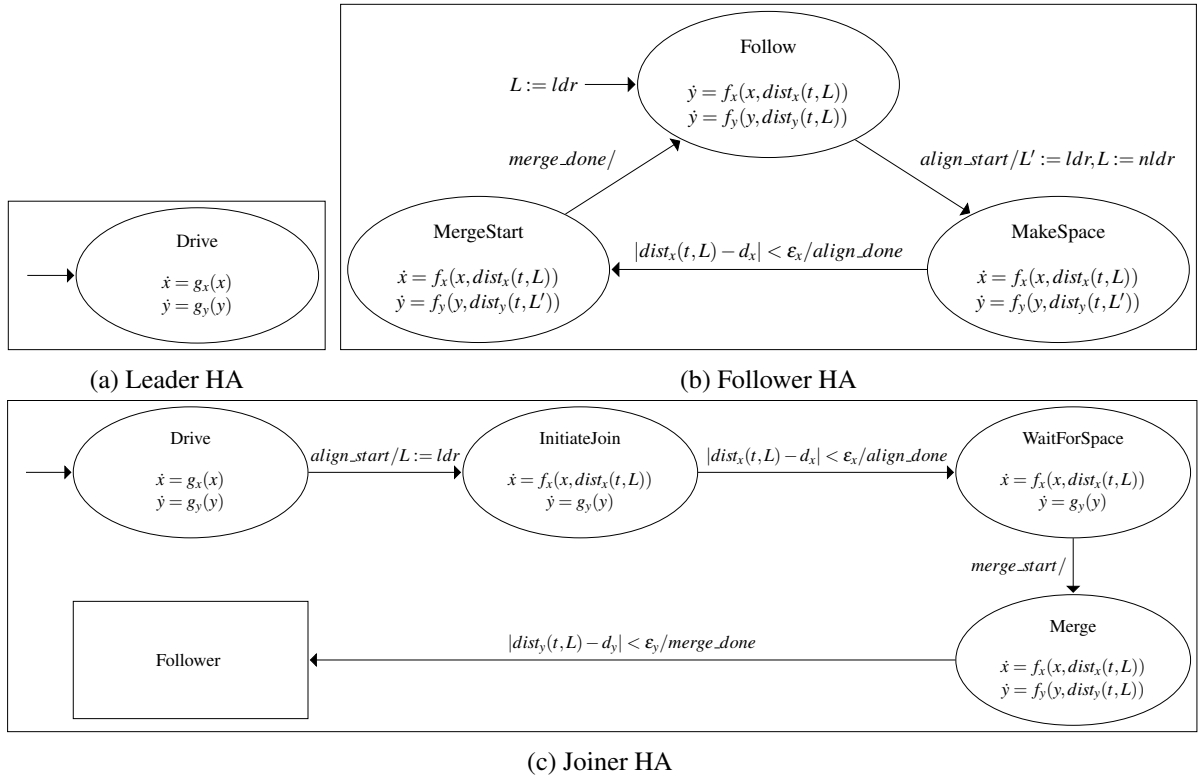
\begin{figure*}
	\centering
	\begin{subfigure}[t]{.28\linewidth}
		\centering
		\resizebox{0.97\linewidth}{!}{\input{"figures/lead_ha.tex"}}
		\caption{Leader HA}
		\label{fig:leader_ha}
	\end{subfigure}%
	\begin{subfigure}[t]{.72\linewidth}
		\centering
		\resizebox{1.0\linewidth}{!}{\input{"figures/follow_ha.tex"}}
		\caption{Follower HA}
		\label{fig:follower_ha}
	\end{subfigure}

	\begin{subfigure}[t]{1.0\linewidth}
		\centering
		\resizebox{1.0\linewidth}{!}{\input{"figures/join_ha.tex"}}
		\caption{Joiner HA}
		\label{fig:join_ha}
	\end{subfigure}
	\caption{Hybrid automata for the low-level layer. Each automata receives input signals from the high-level layer and sends output signals. (a) No inputs/outputs. (b) Inputs: $align\_start$, $merge\_done$ Outputs: $align\_done$ (c) Inputs: $align\_start$, $merge\_start$ Outputs: $align\_done$, $merge\_done$}
	\label{fig:ha_diagram}
\end{figure*}

The low\-/level layer can be modeled as a series of hybrid automata~\cite{alur2015principles} in order to describe the behavior of the controller and plant from Fig. \ref{fig:org_diagram}. Each behavior -- Leader, Follower, and Joiner -- has its own hybrid automata as shown in Fig.~\ref{fig:ha_diagram}. Interaction with the high\-/level layer takes place in the form of input and output signals that directly correspond to unobservable actions provided by the interface in Section~\ref{sec:methodology}. We now formally define these signals with respect to the hybrid automata. Let $\mathbb{N}$ denote the set of natural numbers, then we define:
\begin{align*}
S1: \{align\_start, align\_done\} &\rightarrow \{absent, present\}\\
S2: \{merge\_start, merge\_done\} &\rightarrow \{absent, present\}\\
S3: \{L, L'\} &\rightarrow \mathbb{N}
\end{align*}
Additionally, the following functions are defined. Let the $x$-axis lie parallel to the front of the vehicle and the $y$-axis lie perpendicular to that.
\begin{itemize}
	\item $g_y(y)/g_x(x)$ computes $\dot{y}/\dot{x}$ given the current $y/x$ position while taking into account road geometry. In most cases, this is equivalent to driving forward at a constant velocity.
	\item $f_y(y,\alpha)/f_x(x,\alpha)$ computes $\dot{y}/\dot{x}$ given the current $y/x$ position and a distance $\alpha$ with the objective of reducing $\alpha$ to the minimum safe following distance $d$.
	\item $dist_x(t,L)$ returns the distance along the $x$-axis to vehicle $L$ at time $t$.
	\item $dist_y(t,L)$ returns the distance along the $y$-axis to the center of the lane currently occupied by vehicle $L$ at time $t$. Currently platoons are not allowed to switch lanes so that discontinuities in this distance are avoided.
\end{itemize}
With these signals and functions, we can define each of the vehicle
behaviors.

\subsubsection{Leader}

The Leader automata simply remains in a drive state. No transitions are supported and no input or output signals are accepted.

\subsubsection{Follower}

The Follower automata requires the input signal $align\_start$ to create longitudinal space in front of the vehicle when a merge has been requested. This is accomplished by a change in leader $L$, which coincides with a $set\_ldr$ call in the high\-/level layer. However, only $f_y$ respects the change in leader; $f_x$ continues to be a function of the previous leader, $L'$. This is so the vehicle does not sway in its lane attempting to position itself laterally behind a vehicle that is changing lanes. When the $merge\_done$ signal is received, $L$ is respected on both axes.

\subsubsection{Joiner}

Similarly, the Joiner automata requires the input signal $align\_start$ to position itself behind the vehicle it will be following once the merge is completed. However, the $x$-axis does not respect its new leader, $L$, until the $merge\_start$ signal is received -- indicating that the Follower vehicle is finished aligning itself to this vehicle.

In practice, these automata are implemented as digital controllers that receive input signals from the high\-/level layer via an interface as described in Section~\ref{ssec:highlevel}. The advantage of this setup is that the controllers and vehicle dynamics can be swapped out for any implementation as long as it fulfills the required interface. As an illustrative example, consider a scenario in which the high\-/level behavior described in the previous section has been adopted as an automobile industry standard. Each vehicle manufacturer can then supply a low\-/level layer for controlling its vehicle without having to re\-/verify the behavior protocol.

\section{Experiments} \label{sec:experiments}

\begin{figure}
	\centering
	\begin{subfigure}{.25\linewidth}
		\centering
		\includegraphics[width=1.0\linewidth,angle=90]{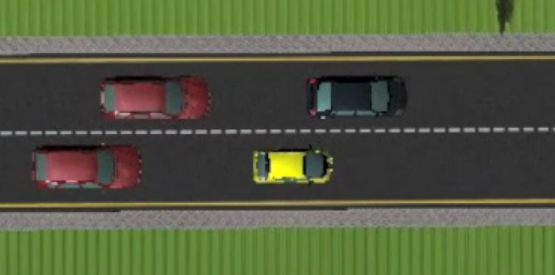}
		\caption{Initial state}
		\label{fig:hetero_convoy_sim_1}
	\end{subfigure}%
	\begin{subfigure}{.25\linewidth}
		\centering
		\includegraphics[width=1.0\linewidth,angle=90]{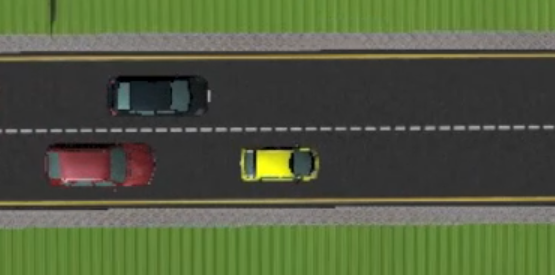}
		\caption{Joiner in position}
		\label{fig:hetero_convoy_sim_2}
	\end{subfigure}%
	\begin{subfigure}{.25\linewidth}
		\centering
		\includegraphics[width=1.0\linewidth,angle=90]{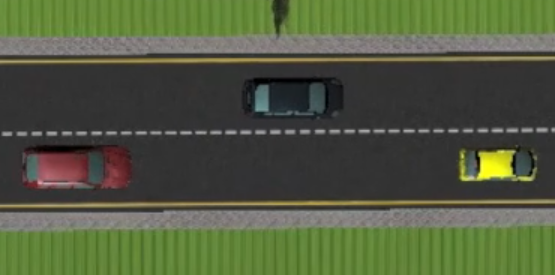}
		\caption{Distance created}
		\label{fig:hetero_convoy_sim_3}
	\end{subfigure}%
	\begin{subfigure}{.25\linewidth}
		\centering
		\includegraphics[width=1.0\linewidth,angle=90]{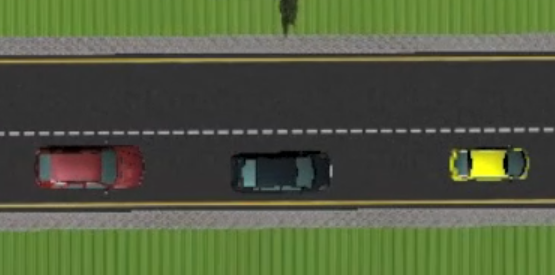}
		\caption{Merge complete}
		\label{fig:hetero_convoy_sim_4}
	\end{subfigure}
	\caption{Actual merge behavior of vehicle platoon.}
	\label{fig:hetero_convoy_sim}
\end{figure}

Simulation is an important step in the analysis and verification of autonomous vehicle platoons, as the closed\-/loop dynamics are a fundamental part of the system. Without simulation, the dynamics are not validated, and much of the problem is left unsolved. To that end, we have developed a simulation framework which ties together the high\-/level and low\-/level layers discussed in this paper and places them in control of automobiles in the Webots~8.3 robotics simulator~\cite{Webots}. Two types of instances were tested: a homogeneous platoon consisting of vehicles with identical physical characteristics and controllers, and a heterogeneous platoon with varying physical characteristics and differing controllers. The behavior of vehicles in these simulations is shown in Fig.~\ref{fig:hetero_convoy_sim}.

\subsection{Homogeneous Platoon}
In the first experimental simulation scenario, four independent vehicles attempted to join a five\-/vehicle platoon. All automobiles were modeled after the Toyota Prius (model included in Webots) with identical physical characteristics and identical manually tuned PID controllers. The controllers generate throttle and brake inputs for the plant (simulated vehicle) and track a target following distance from the vehicle which is being followed. The controllers are not meant to be robust nor optimal, rather they are a simple example of how this framework can be used to control continuous systems. Localization is achieved by means of each vehicle broadcasting out its current position at every simulation step. While not practical in the real world, this is adequate for this simulation as accurately identifying other vehicles and their positions is a challenging task. The distance functions, $dist_x$ and $dist_y$, make use of these broadcasted positions in addition to a forward-facing laser sensor for redundancy. All vehicles eventually successfully merge into the platoon in a distributed and decentralized manner.

\subsection{Heterogeneous Platoon}
The second experimental scenario again consisted of four independent vehicles attempting to join a five\-/vehicle platoon; however, this time the vehicles were of varying models (included in Webots). Four automobiles were modeled after the BMW X5, three after the Citroen C\-/Zero, and two after the Toyota Prius. Additionally, three of the vehicles used the PID controller from the homogeneous platoon experiment while the remaining six used a model predictive controller designed for such scenarios with some minor modifications~\cite{liu2015predictive}. We vary the reference velocity based on the current tracking error, which is similar to previous work~\cite{godbole1994longitudinal}. In addition we modify the cost parameters, Q and R of the controller, so that the weight given to the reference velocity and distance varies based on the error. Once again, all vehicles successfully merged into the platoon without incident.

However, similar to the PID controller in the first experimental simulation, this model predictive controller is not meant to be optimal nor even provide guarantees of stability. Rather its primary purpose is to show that our framework can not only effectively represent interactions between heterogeneous vehicles, but that those vehicles can use dramatically different control schemes.

\section{Conclusions}
\label{sec:conclusions}

In this paper, we have introduced a new framework which utilizes $\pi$-calculus to model complex vehicular networks. This framework allows for the effective modeling of both high\-/level decision\-/making protocols and low\-/level vehicle dynamics, laying the groundwork for future verification of these systems. In order to demonstrate the utility of this framework, we have modeled a decentralized platooning protocol with heterogeneous vehicles and controllers. In future work, we will formally verify complex systems modeled with this framework.



\bibliographystyle{eptcs}
\bibliography{references}

\end{document}

%% file: figures/convoy_mobility1.tex
\tikzset{elliptic state/.style={draw, ellipse, minimum height=2.5cm, minimum width=5cm}}

\fbox{\begin{tikzpicture}[initial text={$L:=v$}, initial distance=1cm, every initial by arrow/.style={-{Stealth[scale=1.5]}}, node distance=2cm, on grid, auto]
\node[state, align=center] (A) {
	A
};

\node[state, align=center] (B) [below = of A] {
	B
};

\node[state, align=center] (C) [below = of B] {
	C
};

\node[state, align=center] (D) [left = of B] {
	D
};

\path[-]
(D) edge node {$x$} (B)
	edge node {$x$} (C) {};
\end{tikzpicture}}

%% file: figures/convoy_mobility2.tex
\tikzset{elliptic state/.style={draw, ellipse, minimum height=2.5cm, minimum width=5cm}}

\fbox{\begin{tikzpicture}[initial text={$L:=v$}, initial distance=1cm, every initial by arrow/.style={-{Stealth[scale=1.5]}}, node distance=2cm, on grid, auto]
\node[state, align=center] (A) {
	A
};

\node[state, align=center] (B) [below = of A] {
	B
};

\node[state, align=center] (C) [below = of B] {
	C
};

\node[state, align=center] (D) [left = of B] {
	D
};

\path[-]
(D) edge node {$y_1$} (B)
	edge node {$y_2$} (C) {};
\end{tikzpicture}}

%% file: figures/convoy_mobility3.tex
\tikzset{elliptic state/.style={draw, ellipse, minimum height=2.5cm, minimum width=5cm}}

\fbox{\begin{tikzpicture}[initial text={$L:=v$}, initial distance=1cm, every initial by arrow/.style={-{Stealth[scale=1.5]}}, node distance=2cm, on grid, auto]
\node[state, align=center] (A) {
	A
};

\node[state, align=center] (B) [below = of A] {
	B
};

\node[state, align=center] (C) [below = of B] {
	C
};

\node[state, align=center] (D) [left = of B] {
	D
};

\path[-]
(D) edge node {$y_2$} (C) {};
\end{tikzpicture}}

%% file: figures/lead_ha.tex
\tikzset{elliptic state/.style={draw, ellipse, minimum height=2.5cm, minimum width=5cm}}

\begin{tikzpicture}[initial text=, initial distance=1cm, every initial by arrow/.style={-{triangle 90}}, shorten >=1pt, node distance=5cm, on grid, auto]
\node[elliptic state, initial, align=center] (q_0) {
	Drive \\[1em] 
	$\dot{x} = g_x(x)$ \\
	$\dot{y} = g_y(y)$
};

\node[fit=(q_0)(current bounding box.west), draw](f_0) {};
\end{tikzpicture}

%% file: figures/follow_ha.tex
\tikzset{elliptic state/.style={draw, ellipse, minimum height=2.5cm, minimum width=5cm}}

\begin{tikzpicture}[initial text={$L:=ldr$}, initial distance=1cm, every initial by arrow/.style={-{triangle 90}}, shorten >=1pt, node distance=5cm, on grid, auto]
\node[elliptic state, initial, align=center] (q_0) {
	Follow \\[1em] 
	$\dot{y} = f_x(x, dist_x(t,L))$ \\
	$\dot{y} = f_y(y, dist_y(t,L))$
};

\node[elliptic state, align=center] (q_1) [below right = of q_0, xshift=2cm] {
	MakeSpace \\[1em] 
	$\dot{x} = f_x(x, dist_x(t,L))$ \\
	$\dot{y} = f_y(y, dist_y(t,L'))$
};


\node[elliptic state, align=center](q_2) [below left = of q_0, xshift=-2cm] {
	MergeStart \\[1em] 
	$\dot{x} = f_x(x, dist_x(t,L))$ \\
	$\dot{y} = f_y(y, dist_y(t,L'))$
};

\path[-{triangle 90}]
(q_0) edge node {$align\_start / L':=ldr, L:=nldr$} (q_1)
(q_1) edge node [midway, above]{$|dist_x(t,L)-d_x| < \epsilon_x / align\_done$} (q_2)
(q_2) edge node {$merge\_done/$} (q_0);
\node[fit=(q_0)(q_1)(q_2), draw](f_0) {};
\end{tikzpicture}

%% file: figures/join_ha.tex
\tikzset{elliptic state/.style={draw, ellipse, minimum height=2.5cm, minimum width=5cm}}
\tikzset{rectangular state/.style={draw, rectangle, minimum height=2.5cm, minimum width=5cm}}

\begin{tikzpicture}[initial text={}, initial distance=1cm, every initial by arrow/.style={-{triangle 90}}, shorten >=1pt, node distance=4cm, on grid, auto]
\node[elliptic state, initial, align=center] (q_-1) {
	Drive \\[1em] 
	$\dot{x} = g_x(x)$ \\
	$\dot{y} = g_y(y)$
};

\node[elliptic state, align=center] (q_0) [right = of q_-1, xshift=4.3cm] {
	InitiateJoin \\[1em] 
	$\dot{x} = f_x(x, dist_x(t,L))$ \\
	$\dot{y} = g_y(y)$
};

\node[elliptic state, align=center] (q_1) [right = of q_0, xshift=6.5cm] {
	WaitForSpace \\[1em] 
	$\dot{x} = f_x(x, dist_x(t,L))$ \\
	$\dot{y} = g_y(y)$
};

\node[elliptic state, align=center] (q_2) [below = of q_1] {
	Merge \\[1em] 
	$\dot{x} = f_x(x, dist_x(t,L))$ \\
	$\dot{y} = f_y(y, dist_y(t,L))$
};

\node[rectangular state, align=center] (q_3) [below = of q_-1] {
	Follower
};

\path[-{triangle 90}]
(q_-1) edge node {$align\_start / L:=ldr$} (q_0)
(q_0) edge node {$|dist_x(t,L)-d_x| < \epsilon_x / align\_done$} (q_1)
(q_1) edge node [midway, left] {$merge\_start/$} (q_2)
(q_2) edge node [midway, above] {$|dist_y(t,L)-d_y| < \epsilon_y / merge\_done$} (q_3);


\node[fit=(q_-1)(q_0)(q_1)(q_2)(q_3)(current bounding box.west), draw](f_0) {};
\end{tikzpicture}